\documentclass[11pt,a4paper]{article}
\usepackage{times,latexsym}
\usepackage{url}
\usepackage[T1]{fontenc}
\usepackage{array}
\usepackage{CJKutf8}
\usepackage{tcolorbox}
\usepackage{amssymb}
\usepackage{makecell}
\usepackage{latexsym}
\usepackage{enumitem}
\usepackage{amsmath}
\usepackage{subfigure}
\usepackage{multirow}
\usepackage{booktabs}
\usepackage{graphicx}

\newcommand{\literary}{\textcolor{purple}}
\newcommand{\cometkiwi}{\textcolor{black}}
\newcommand{\modi}{\textcolor{black}}

\def\huggingface{\raisebox{-1.5pt}{\includegraphics[height=0.90em]{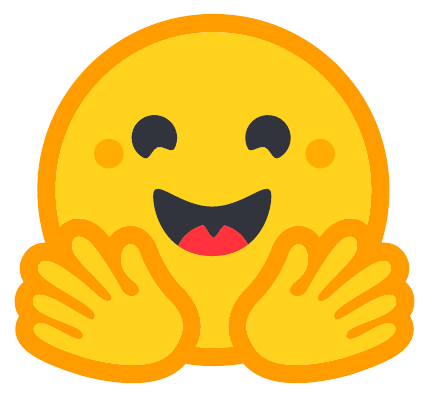}}}

\usepackage[acceptedWithA]{tacl2021v1}

\usepackage{xspace,mfirstuc,tabulary}

\newif\iftaclinstructions
\taclinstructionsfalse 
\iftaclinstructions
\renewcommand{\confidential}{}
\renewcommand{\anonsubtext}{(No author info supplied here, for consistency with
TACL-submission anonymization requirements)}
\newcommand{\instr}
\fi

\iftaclpubformat 

\else

\fi

\title{DeepTrans: Deep Reasoning Translation via Reinforcement Learning}

\author{Jiaan Wang, \ Fandong Meng\thanks{ \ \ Corresponding author.}, \ Jie Zhou \\
Pattern Recognition Center, WeChat AI, Tencent Inc \\ 
\texttt{\{torchwang,fandongmeng,yxuezhang,withtomzhou\}@tencent.com}
}

\date{}

\begin{document}
\maketitle
\begin{abstract}
Recently, deep reasoning LLMs (\emph{e.g.}, OpenAI o1 and DeepSeek-R1) have shown promising performance in various downstream tasks.
Free translation is an important and interesting task in the multilingual world, which requires going beyond word-for-word translation.
However, the task is still under-explored in deep reasoning LLMs.
In this paper, we introduce \textbf{DeepTrans}, a deep reasoning translation model that learns free translation via reinforcement learning (RL).
Specifically, we carefully build a reward model with pre-defined scoring criteria on both the translation results and the thought processes.
The reward model teaches DeepTrans how to think and free-translate the given sentences during RL.
Besides, our RL training does not need any labeled translations, avoiding the human-intensive annotation or resource-intensive data synthesis.
Experimental results show the effectiveness of DeepTrans.
Using Qwen2.5-7B as the backbone, DeepTrans improves performance by 16.3\% in literature translation, and outperforms strong deep reasoning LLMs.
Moreover, we summarize the failures and interesting findings during our RL exploration.
We hope this work could inspire other researchers in free translation.\footnote{\url{https://github.com/krystalan/DRT}}
\end{abstract}

\section{Introduction}

Recently, deep reasoning LLMs~\cite{openai_o1_2024,guo2025deepseek} have shown remarkable performance in tasks ranging from math~\cite{chen2025towards,li2025system}, \modi{to coding}~\cite{zhang2024o1}, question-taking~\cite{guan2025deeprag}, etc.

Some researchers bring the success of deep reasoning LLMs to neural machine translation (MT)\modi{, with a particular focus on free translation rather than word-for-word translation~\cite{zhao2024marco,wang2024drt,liu2025new}.
Free translation is a translation approach that allows more flexibility to adapt the text to the target language, \emph{taking into account} cultural nuances and making the text more natural and understandable for the target audience~\cite{barbe1996dichotomy,chen-etal-2018-detecting}.
Given that, free translation is well-suited for deep reasoning LLMs to perform.}
Marco-o1~\cite{zhao2024marco} \modi{aims to extend deep reasoning LLMs into general domains where clear standards may be lacking.} It uses illustrative examples to show the effectiveness of long chain-of-thought (CoT) in colloquial and slang MT.
\citet{wang2024drt} further systematically study the long CoT in literature MT.
They find the literature sentences with metaphors or similes generally require cultural background to understand.
Based on this insight, \citet{wang2024drt} propose \modi{deep reasoning translation (DRT)} models to translate the literature text from English to Chinese with step-by-step reasoning.
The models are trained from synthesized long CoT translation data.
\citet{chen2025evaluating} conduct empirical studies on how deep reasoning LLMs work on MT.
They verify the strengths of deep reasoning LLMs in historical and cultural translation, but also point out their issues, \emph{e.g.}, LLMs do not follow the instruction and fail to translate.
More recently, R1-T1~\cite{he2025r1} is presented, which is the first attempt to employ reinforcement learning (RL) in deep reasoning LLMs.
Specifically, it uses COMET~\cite{rei-etal-2020-comet} score as the reward signal to optimize MT LLMs via modified REINFORCE++ (a RL training algorithm)~\cite{hu2025reinforce++}.
MT-R1~\cite{feng2025mt} uses a combination of BLEU and CometKiwi as the reward signal to optimize MT LLMs, and adopts GRPO~\cite{shao2024deepseekmath} as the RL algorithm.

Meanwhile, RL has been verified to have a strong ability in deep reasoning LLMs, and LLMs can be equipped with powerful reasoning ability solely based on RL~\cite{guo2025deepseek,yu2025dapo}.
However, RL is still under-explored in deep reasoning MT.
In detail, Marco-o1 and DRT are only trained via supervised fine-tuning (SFT).
Though R1-T1 and MT-R1 adopt RL training, using the COMET and CometKiwi as the rewards is not well calibrated.
As pointed out by \citet{liu2025new}, deep reasoning MT LLMs might achieve better translation but a lower COMET.
Besides, in literature domains, COMET and CometKiwi lose their effectiveness as evaluation metrics and show a poor correlation with \modi{human judgement}~\cite{karpinska-iyyer-2023-large,wang2024drt}.

In this paper, our research goal is to explore \emph{how to improve the free translation ability of deep reasoning LLMs via reinforcement learning}.
Following~\citet{wang2024drt}, our explorations focus on the literature domain, where free translation is essential for addressing cultural differences.
Subsequently, we focus on the reward modeling, and study \emph{how to employ an effective reward model during RL training}.
There \modi{are} three mainstream types of reward models in previous work:
\romannumeral1) using MT metrics such as BLEU~\cite{papineni-etal-2002-bleu}, COMET~\cite{rei-etal-2020-comet} and CometKiwi~\cite{rei-etal-2022-cometkiwi};
\romannumeral2) training a reward model with preference data~\cite{stiennon2020learning,nakano2021webgpt}, and using the reward model to \modi{infer} a scalar reward during RL training;
\romannumeral3) designing rule-based rewards~\cite{guo2025deepseek}.
However, these rewards might lose their effectiveness or require high-quality annotated data in MT, hindering their usage.
In detail, for \romannumeral1), the reference-based BLEU and COMET are unsuitable for literature MT since the high-quality translation data is difficult to collect.
Though CometKiwi is a reference-free metric, its effectiveness is limited in the literature domain~\cite{wang2024drt}.
For \romannumeral2), training a reward model requires large-scale preference data, which is difficult to collect due to the expensive annotation costs.
For \romannumeral3), the rule-based rewards are suitable for tasks whose answers are easy to verify, \emph{e.g.}, math and code problems~\cite{swamy2025all}.
In MT, we cannot design simple rules to judge the quality of translations.

In view of the strong ability of LLM-as-a-judge~\cite{wang-etal-2023-chatgpt,kocmi-federmann-2023-large,li2024generation}, we decide to use an advanced LLM, \emph{i.e.}, DeepSeek-v3 (671B)~\cite{liu2024deepseek}, as the reward model.
Specifically, we carefully design pre-defined scoring criteria on both the translation results and the thought processes.
For the generation of deep reasoning MT LLMs, the reward model takes the pre-defined criteria into account to provide a discrete reward (\emph{e.g.}, 3-point or 100-point scores).
In this manner, we avoid collecting human-annotated preference data for training the reward model.
In addition, the effectiveness of the reward model could also be ensured when we adopt the state-of-the-art LLM.
We conduct extensive experiments on literature MT, using the above RL strategy to train DeepTrans-7B (with the backbone of Qwen2.5-7B).
Experiments show the effectiveness of our method, DeepTrans-7B improves performance by 16.3\% and outperforms strong deep reasoning baselines (\emph{e.g.}, QwQ-32B-preview).
Besides, only training with source sentences, DeepTrans-7B outperforms DRT-7B (which is trained on synthesized long thought and translation data) in terms of both automatic metrics and GPT-4o evaluation.
Moreover, we share the failures in our pilot RL experiments and summarize the critical component in reward modeling.

Our contributions are as follows:
\begin{itemize}[leftmargin=*,topsep=0pt]
\setlength{\itemsep}{0pt}
\setlength{\parsep}{0pt}
\setlength{\parskip}{0pt}
\item We propose DeepTrans, which aims to enhance the free translation ability of deep reasoning LLMs via RL. In detail, we use an LLM as the reward model, and carefully design scoring criteria on both translations and thought processes.
\item Experimental results verify the effectiveness of our DeepTrans. Only training with the source sentences, DeepTrans outperforms strong deep reasoning baselines and LLMs that are fine-tuned with synthesized MT data.
\item We summarize the failures and critical components during the RL training to provide a deeper understanding of deep reasoning MT LLMs.
\end{itemize}

\section{Related Work}

\noindent \textbf{Deep Reasoning LLMs.}
In recent years, deep reasoning LLMs have pioneered a growing research in long chain-of-thought (CoT) reasoning.
\modi{Different from the non-reasoning-oriented LLMs, deep reasoning LLMs involve a more detailed, iterative process of exploration and reflection within a given problem space~\cite{chen2025towards,li2025survey}.
Many studies investigate the mathematical reasoning, programming tasks, and multidisciplinary knowledge reasoning capabilities of deep reasoning LLMs, and achieve promising performance~\cite{yu2024natural,sun2025survey,yax2024studying,li2025system,guan2025deeprag,jin2025search}.}
Some researchers investigate the MT capability of deep reasoning LLMs.
\citet{zhao2024marco} and \citet{liu2025new} briefly show that the long CoT reasoning helps the model to reach more idiomatic translations.
DRT~\cite{wang2024drt} is further proposed in literary translation, and it is trained with synthesized SFT data.
More recently, R1-T1~\cite{he2025r1} and MT-R1~\cite{feng2025mt} employ RL to improve the translation ability of deep reasoning LLMs.
\modi{However, R1-T1 and MT-R1 use the traditional MT metrics, \emph{i.e.}, BLEU~\cite{papineni-etal-2002-bleu}, COMET~\cite{rei-etal-2020-comet} and COMETKiwi~\cite{rei-etal-2022-cometkiwi}, as the RL reward.
Different from them, we employ an advanced LLM with pre-defined scoring criteria to qualify the quality of both the translation and the thought process as the reward.
In this way, the flaws of traditional metrics~\cite{karpinska-iyyer-2023-large,wang2024drt} can be avoided, and the strong ability of LLM-as-a-judge can be utilized to guide the RL training process effectively.}

\noindent \textbf{RL in Traditional MT.}
Machine translation via reinforcement learning has been explored before the deep reasoning LLM era. Early work tries to train MT models via optimizing BLEU scores~\cite{ranzato2015sequence,shen-etal-2016-minimum,bahdanau2017actor}.
\citet{wu2016google} design GLEU scores as the rewards to deal with the drawbacks of BLEU in single sentence evaluation.
\citet{wu-etal-2018-study} propose a method to involve large-scale monolingual data during RL training.
\citet{choshenweaknesses} show the challenges of optimizing MT models via RL, \emph{e.g.}, sparse reward signals and high-dimensional action space.
\citet{kiegeland-kreutzer-2021-revisiting} provide further analyses on these challenges.
\citet{kang-etal-2020-dynamic} study document-level MT, and they propose a new method to both select context and translate sentences via RL.

\section{DeepTrans}
\begin{figure*}[t]
\centerline{\includegraphics[width=0.95\textwidth]{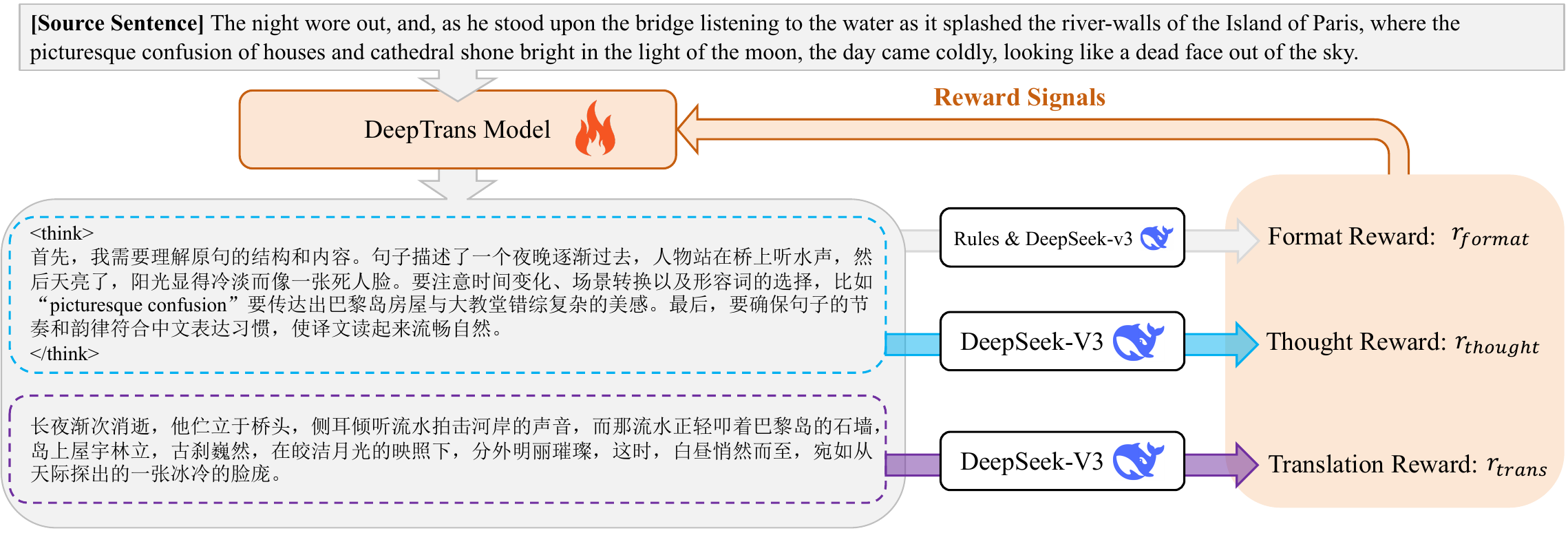}}
\caption{The overview of DeepTrans RL training.}
\label{fig:rl_framework}
\end{figure*}

In this section, we introduce DeepTrans. As illustrated in Figure~\ref{fig:rl_framework}, there are three types of rewards: \emph{format reward}, \emph{thought reward} and \emph{translation reward}.
We first discuss the rewards designed in the RL framework (\S~\ref{subsec:2.1}) and then provide the training details of DeepTrans (\S~\ref{subsec:2.2}).

\subsection{Reward Modeling}
\label{subsec:2.1}

Given a source sentence, DeepTrans first thinks about how to translate the sentence, and then provides the translation result.
We design the format of model generation as ``\texttt{<think>} [thought] \texttt{</think>} [translation]'', where ``\texttt{<think>}'' and ``\texttt{</think>}'' are two special tokens to indicate the boundary of thought content. ``[thought]'' and ``[translation]'' denote the content of thought and translation, respectively.
Based on the model generation, we design the following rewards:

\vspace{0.5ex}
\noindent \emph{\textbf{Format Reward}}: We use a regular expression to judge whether the generation format is correct. Besides, in pilot experiments, we find that there might be some explanations in the translation results. To avoid it, we employ DeepSeek-v3~\cite{liu2024deepseek} to judge whether the translation results only contain translations. The judgment prompt is shown as follows:

\begin{tcolorbox}
\fontsize{10pt}{10pt}\selectfont

A translation question requires translating a given text from [src lang] into [trg lang]. \\

The given text is as follows:\\
<text>\\
\{src\}\\
</text>\\

Someone did this translation task and the translation result is as follows:\\
<translation>\\
\{trans\}\\
</translation>\\

Please judge whether the translation result belongs to the following situations:\\
1. It contains only the translation result.\\
2. It contains the translation result and the explanation.\\
3. It does not contain the translation result, but only the explanation.\\

Please directly output your judgment result, such as: ``Judgment result: 1'', ``Judgment result: 2'' or ``Judgment result: 3''

\end{tcolorbox}
\noindent where ``\{src\}'' and ``\{trans\}'' denote the source sentence and the translation result, respectively.

If both (a) the generation format is correct (determined by the regular expression), and (b) the translation result does not contain any explanations (determined by DeepSeek-v3), we regard the format as correct; otherwise, it is incorrect:
\begin{equation}
\small
r_{\text{format}} = 
\begin{cases} 
1 & \text{if format is correct} \\
0 & \text{if format is incorrect}
\end{cases}
\end{equation}

\noindent \emph{\textbf{Thought Reward}}: The thought process is important to guide the final translation.
To reward meaningful thought processes, we adopt DeepSeek-v3 to provide the thought reward with a 3-point scale:

\begin{tcolorbox}
\fontsize{10pt}{10pt}\selectfont

A translation question requires translating a given text from [src lang] into [trg lang]. \\

The given text is as follows:\\
<text>\\
\{src\}\\
</text>\\

Someone did this translation question, and began to think how to translate:\\
<think>\\
\{think\}\\
</think>\\

Please judge whether there is a detailed analysis of the given text in this thinking process:\\
1. A lack of analysis: Only very shallow thinking was done, and no detailed analysis of the given text was carried out.\\
2. Slight analysis: The given text was analyzed in detail, and how to translate it was discussed in detail.\\
3. Detailed analysis: The given text was analyzed in detail, and various translation possibilities were discussed in detail, and trade-offs were made.\\

Please directly output your judgment results, such as: ``a lack of analysis'', ``slight analysis'' or ``detailed analysis''

\end{tcolorbox}
\noindent where ``\{think\}'' denotes the thought content. Subsequently, we define the thought reward as:
\begin{equation}
\small 
  r_{\text{thought}} = 
\begin{cases} 
2 & \text{if} \operatorname{v3}^{\text{th}}(\text{src}, \text{think}) = \text{detailed analysis} \\
1 & \text{if} \operatorname{v3}^{\text{th}}(\text{src}, \text{think}) = \text{slight analysis} \\
0 & \text{if} \operatorname{v3}^{\text{th}}(\text{src}, \text{think}) = \text{a lack of analysis} \\
\end{cases}
\end{equation}
where $\operatorname{v3}^{\text{th}}(\cdot,\cdot)$ denotes using DeepSeek-v3 as the thought reward scorer.

\vspace{0.5ex}
\noindent \emph{\textbf{Translation Reward}}:
We also employ DeepSeek-v3 to assess the translation quality. We borrow the prompt from \citet{wang2024drt}, which is designed to evaluate the quality of literature translation:

\begin{tcolorbox}
\fontsize{10pt}{10pt}\selectfont

\texttt{SYSTEM PROMPT:} \\

Please evaluate the following Chinese translation of an English text. Rate the translation on a scale of 0 to 100, where:\\
- 10 points: Poor translation; the text is somewhat understandable but contains significant errors and awkward phrasing that greatly hinder comprehension for a Chinese reader.\\
- 30 points: Fair translation; the text conveys the basic meaning but lacks fluency and contains several awkward phrases or inaccuracies, making it challenging for a Chinese reader to fully grasp the intended message.\\
- 50 points: Good translation; the text is mostly fluent and conveys the original meaning well, but may have minor awkwardness or slight inaccuracies that could confuse a Chinese reader.\\
- 70 points: Very good translation; the text is smooth and natural, effectively conveying the intended meaning, but may still have minor issues that could slightly affect understanding for a Chinese reader.\\
- 90 points: Excellent translation; the text is fluent and natural, conveying the original meaning clearly and effectively, with no significant issues that would hinder understanding for a Chinese reader.\\

Please provide the reason first, followed by a score. Format your evaluation in the JSON structure below:\\
\{"reason": "reason for the score", "score": int\}\\

\end{tcolorbox}

\begin{tcolorbox}
\fontsize{10pt}{10pt}\selectfont
\texttt{USER PROMPT:} \\

<text>\\
\{src\}\\
</text>\\
<translation>\\
\{trans\}\\
</translation>

\end{tcolorbox}

In this way, DeepSeek-v3 will generate its judgment for the translation result and provide a reward score on a 100-point scale:
\begin{equation}\label{eq:3}
\small
r_{\text{trans}} = \operatorname{v3}^{\text{tr}}(\text{src}, \text{trans})
\end{equation}
where $\operatorname{v3}^{\text{tr}}(\cdot,\cdot)$ denotes using DeepSeek-v3 as the translation reward scorer.

\vspace{0.5ex}
\noindent \emph{\textbf{Overall Reward}}: Given the above three types of rewards, we finally design the overall reward:
\begin{equation}\label{eq:4}
\small
  r_{\text{all}} = 
\begin{cases} 
0 & \text{if  } r_{\text{format}} = 0 \\
r_{\text{trans}} + \alpha \times r_{\text{thought}} & \text{if  } r_{\text{format}} \neq 0
\end{cases}
\end{equation}
where $\alpha$ is a trade-off hyperparameter between $r_{\text{trans}}$ and $r_{\text{thought}}$.
Note that DeepSeek-v3 is used in $r_\text{trans}$ / $r_\text{thought}$, and its effectiveness in MT evaluation is also verified by~\citet{deepseekv3mteval}.

\subsection{Training Details}
\label{subsec:2.2}

\noindent \emph{\textbf{Cold Start SFT.}}
We use a non-reasoning LLM, \emph{i.e.}, Qwen2.5-7B-Instruct~\cite{yang2024qwen2}, as the backbone of DeepTrans.
To adapt DeepTrans to the deep reasoning MT, we first use DeepSeek-R1~\cite{guo2025deepseek} to generate seed translation samples in the general domain (\emph{instead of the literature domain}).
These samples follow our designed format, and are used to few-shot SFT the backbone model (named cold start SFT).
The goal of cold start SFT is not to teach the model free translation, but the general translation with long thought.

\noindent \emph{\textbf{RL Training.}}
In view of the strong ability of GRPO~\cite{shao2024deepseekmath}, we adopt it in our RL training.
For the policy model $\pi$, given it a source sentence $s$, GRPO first samples a number of generations $\{g_1, g_2, ..., g_n\}$ based on $\pi$, where each $g_i$ involves a thought process and the translation result.
Then, GRPO optimizes the policy model $\pi'$ by maximizing the following objective:
\begin{equation}
\small
\frac{1}{n}\sum_{1}^{n}(\operatorname{min}(\nabla_{\pi}A_i, \operatorname{clip}(\nabla_{\pi}, 1-\epsilon, 1 + \epsilon)A_i) -\beta \mathcal{D}
\end{equation}
\begin{equation}
\small
\nabla_{\pi} = \frac{\pi'(g_i|s)}{\pi(g_i|s)}
\end{equation}
\begin{equation}
\small
\mathcal{D} = \mathbb{D}_{kl}(\pi' || \pi_{ref}))
\end{equation}
where $\epsilon$ and $\beta$ are hyperparameters. $\pi_{ref}$ is the reference model, and $\mathbb{D}_{kl}(\pi' || \pi_{ref}))$ indicates the KL divergence between $\pi'$ and $\pi_{ref}$.
$A_i$ denotes the advantage that is calculated as follows:
\begin{equation}
\small
A_i = \frac{r_{\text{all}}^{i} - {\operatorname{mean}(\{r_{\text{all}}^1, r_{\text{all}}^2, \cdots, r_{\text{all}}^n\})}}{{\operatorname{std}(\{r_{\text{all}}^1, r_{\text{all}}^2, \cdots, r_{\text{all}}^n\})}}.
\end{equation}
where $r_{\text{all}}^i$ denotes the overall reward of $g_i$.

\section{Experiments}

\subsection{Experimental Setups}

\noindent \textbf{Data.}
We use MetaphorTrans~\cite{wang2024drt} in experiments, which is an English-Chinese literature MT data involving 19K training, 1K validation and 2K test samples. Each sample involves a source English sentence, the corresponding Chinese translation, and the thought process during translation.
The source sentences are selected from English literature books, and generally contain metaphors or similes.
The thought processes and translation results are synthesized via Qwen2.5-72B-Instruct~\cite{yang2024qwen2}.
We only use the source sentences during RL training.

In addition to the MetaphorTrans test set, we purchase electronic copies of two complete literature books, and use them to evaluate deep reasoning MT models:
(1) \emph{The Essential O. Henry Collection} (by O. Henry) and (2) \emph{Orbital} (by Samantha Harvey).\footnote{\modi{We will release the test data of \emph{The Essential O. Henry Collection}, and release the data processing script of \emph{Orbital} (the content of \emph{Orbital} cannot be directly released due to copyright protection).}}
Both books are rich in literary nuance, making it challenging for even human translators to achieve a free translation.

\vspace{0.5ex}
\noindent \textbf{Metrics.}
Since (1) the references in MetaphorTrans are synthesized via LLMs, and are not verified by human translators; and (2) the golden references of \emph{The Essential O. Henry Collection} and \emph{Orbital} are missing, we adopt reference-free metrics in our experiments.
Specifically, we use \emph{CometKiwi}~\cite{rei-etal-2022-cometkiwi} to evaluate the model translations, which judges whether a translation conveys the semantics of the source sentence.
Moreover, following \citet{wang2024drt}, we use evaluators implemented using GPT-4o in two reference-free manners, which we refer to as \emph{GRF} and \emph{GEA}, respectively.
The evaluation prompt of \emph{GRF} borrows from \citet{kocmi-federmann-2023-large}.\footnote{Please refer to Figure 1 in \citet{kocmi-federmann-2023-large}}
For \emph{GEA}, the prompt mainly borrows from \citet{wang2024drt}, and we employ two variants of \emph{GEA}, \emph{i.e.}, \emph{GEA100} and \emph{GEA5}.
The evaluation prompt of \emph{GEA100} is the same as the translation reward illustrated in \S~\ref{subsec:2.1}, while that of \emph{GEA5} simply narrows the scoring scope of \emph{GEA100} from a 100-point to a 5-point scale.
Among them, \emph{GRF} evaluates translations from a general perspective while \emph{GEA5} and \emph{GEA100} evaluate translations from a literary perspective.
\modi{The effectiveness of GRF in general translation and GEA in literary translation is demonstrated by \citet{kocmi-federmann-2023-large} and \citet{wang2024drt}, respectively.}
Since \emph{GRF}, \emph{GEA5} and \emph{GEA100} need the costs of OpenAI's API, we randomly select 400 samples from each test set to conduct evaluation.

\vspace{0.5ex}
\noindent \textbf{Backbone.}
Given the high computation costs in RL, we try to use LLMs (< 10B parameters) as the backbone.
Among all LLMs, Qwen2.5-7B~\cite{yang2024qwen2} and Llama3-8B~\cite{grattafiori2024llama} are state-of-the-art choices.
\citet{wang2024drt} show that Qwen2.5-7B outperforms Llama3-8B in literature translation. Thus, we use Qwen2.5-7B as the backbone of DeepTrans.

\begin{table}[t]
\centering
\resizebox{0.45\textwidth}{!}
{
\begin{tabular}{lcccc}
\toprule[1pt]
& \multirow{2}{*}{Para.} & Deep      & Training on & \multirow{2}{*}{Ckpt.} \\
&                             & Reason. & MetaphorTrans    &                        \\ \midrule[1pt]
\multicolumn{5}{c}{\emph{General Non-reasoning Baselines}}                                                           \\
Llama-3.1-8B-Instruct        & 8B                          & $\times$         & $\times$           &     \href{https://huggingface.co/meta-llama/Llama-3.1-8B-Instruct}{\huggingface}                   \\
Qwen2.5-7B-Instruct          & 7B                          & $\times$         & $\times$           &   \href{https://huggingface.co/Qwen/Qwen2.5-7B-Instruct}{\huggingface}                     \\
Qwen2.5-14B-Instruct         & 14B                         & $\times$         & $\times$           &    \href{https://huggingface.co/Qwen/Qwen2.5-14B-Instruct}{\huggingface}                    \\
GPT-4o                       & -                           & $\times$         & $\times$           &         -               \\ \midrule[1pt]
\multicolumn{5}{c}{\emph{General Reasoning Baselines}}                                                               \\
Marco-o1-7B                  & 7B                          & \checkmark         & $\times$           &     \href{https://huggingface.co/AIDC-AI/Marco-o1}{\huggingface}                   \\
QwQ-32B-preview              & 32B                         & \checkmark         & $\times$           &      \href{https://huggingface.co/Qwen/QwQ-32B-Preview}{\huggingface}                  \\
QwQ-32B                      & 32B                         & \checkmark         & $\times$           &     \href{https://huggingface.co/Qwen/QwQ-32B}{\huggingface}                   \\
DeepSeek-Qwen-7B  & 7B                          & \checkmark         & $\times$           &   \href{https://huggingface.co/deepseek-ai/DeepSeek-R1-Distill-Qwen-7B}{\huggingface}                     \\
DeepSeek-Llama-8B & 8B                          & \checkmark         & $\times$           &   \href{https://huggingface.co/deepseek-ai/DeepSeek-R1-Distill-Llama-8B}{\huggingface}                     \\
DeepSeek-Qwen-14B & 14B                         & \checkmark         & $\times$           &   \href{https://huggingface.co/deepseek-ai/DeepSeek-R1-Distill-Qwen-14B}{\huggingface}                     \\
DeepSeek-Qwen-32B & 32B                         & \checkmark         & $\times$           &   \href{https://huggingface.co/deepseek-ai/DeepSeek-R1-Distill-Qwen-32B}{\huggingface}                     \\
DeepSeek-R1                  & 671B                        & \checkmark         & $\times$           &  \href{https://huggingface.co/deepseek-ai/DeepSeek-R1}{\huggingface}                      \\
o1-preview                   & -                           & \checkmark         & $\times$           &          -              \\ \midrule[1pt]
\multicolumn{5}{c}{\emph{MT Non-reasoning Baselines}}                                                                \\
Llama-3.1-8B-SFT             & 8B                          & $\times$         & \checkmark           &       -                 \\
Qwen2.5-7B-SFT               & 7B                          & $\times$         & \checkmark           &       -                 \\
Qwen2.5-14B-SFT              & 14B                         & $\times$         & \checkmark           &       -                 \\ \midrule[1pt]
\multicolumn{5}{c}{\emph{MT \modi{Reasoning} Baselines}}                                                                \\
DRT-7B                       & 7B                          & \checkmark         & \checkmark           &     \href{https://huggingface.co/Krystalan/DRT-7B}{\huggingface}                   \\
DRT-8B                       & 8B                          & \checkmark         & \checkmark           &     \href{https://huggingface.co/Krystalan/DRT-8B}{\huggingface}                   \\
DRT-14B                      & 14B                         & \checkmark         & \checkmark           &     \href{https://huggingface.co/Krystalan/DRT-14B}{\huggingface}                   \\ \midrule[1pt]
\multicolumn{5}{c}{\emph{Our}}                                                                                       \\ 
DeepTrans-7B                 & 7B                          & \checkmark         & \checkmark           &  \\ \bottomrule[1pt]
\end{tabular}
}
\caption{Comparisons between baselines. Para.: Parameter; Reason.: Reasoning. Ckpt.: Checkpoint.}
\label{table:baselines}
\end{table}

\begin{table*}[t]
\centering
\resizebox{0.98\textwidth}{!}
{
\begin{tabular}{lcccccccccccc}
\toprule[1pt]
\multicolumn{1}{c}{\multirow{2}{*}{Model}}      & \multicolumn{4}{c}{MetaphorTrans}                                & \multicolumn{4}{c}{\emph{O. Henry}}                                     & \multicolumn{4}{c}{\emph{Orbital}}                                      \\
\cmidrule(r){2-5} \cmidrule(r){6-9} \cmidrule(r){10-13} & GRF            & GEA5          & GEA100         & CometKiwi      & GRF            & GEA5          & GEA100         & CometKiwi      & GRF            & GEA5          & GEA100         & CometKiwi      \\ \midrule[1pt]
Llama-3.1-8B-Instruct     & 79.25          & 3.31          & 59.58          & 70.14          & 79.73          & 3.43          & 57.17          & 74.35          & 79.92          & 3.54          & 59.90           & 75.09          \\
Qwen2.5-7B-Instruct       & 81.53          & 3.62          & 66.21          & 70.36          & 85.26          & 3.83          & 66.50           & 76.18          & 83.38          & 4.00             & 70.10           & 76.46          \\
Qwen2.5-14B-Instruct      & 84.74          & 3.87          & 70.86          & 72.01          & 86.83          & 3.98          & 70.53          & 77.04          & 84.39          & 4.09          & 71.65          & 76.55          \\
Marco-o1-7B               & 82.41          & 3.57          & 64.24          & 71.62          & 83.12          & 3.71          & 63.11          & 76.00             & 81.84          & 3.89          & 67.64          & 75.38          \\
DeepSeek-Qwen-7B          & 65.16          & 2.67          & 43.66          & 63.49          & 68.97          & 2.86          & 45.64          & 70.67          & 71.28          & 3.16          & 51.91          & 72.43          \\
DeepSeek-Llama-8B         & 76.31          & 3.24          & 56.89          & 67.13          & 78.17          & 3.39          & 56.14          & 73.39          & 78.91          & 3.64          & 59.75          & 74.47          \\
DeepSeek-Qwen-14B         & 83.92          & 3.81          & 70.64          & 71.01          & 83.27          & 3.82          & 64.79          & 75.22          & 82.30           & 4.01          & 69.10           & 76.28          \\ \midrule
Llama-3.1-8B-SFT$^{\diamondsuit}$          & 84.10           & 3.88          & 69.33          & 70.25          & 85.04          & 3.87          & 66.60           & 76.14          & 80.37          & 3.87          & 64.38          & 75.11          \\
Qwen2.5-7B-SFT$^{\diamondsuit}$            & 85.06          & 3.93          & 72.29          & 71.03          & 86.84          & \underline{4.05}          & 71.05          & \underline{77.29}          & 85.46          & 4.12          & 70.55          & 76.32          \\
Qwen2.5-14B-SFT$^{\diamondsuit}$           & 85.66          & 4.02          & 74.53          & \underline{72.08}          & 87.27          & \underline{4.05}          & \underline{73.06}          & \textbf{77.54} & 85.55          & \underline{4.14}          & \underline{75.84}          & \textbf{77.40}           \\
DRT-7B$^{\diamondsuit}$                    & 85.57          & 4.05          & 75.05          & 71.78          & 86.36          & 3.96          & 69.51          & 76.12          & 81.69          & 3.84          & 65.56          & 69.95          \\
DRT-8B$^{\diamondsuit}$                    & 84.49          & 3.91          & 69.65          & 70.85          & 83.61          & 3.75          & 64.76          & 73.89          & 79.14          & 3.65          & 61.36          & 66.36          \\
DRT-14B$^{\diamondsuit}$                   & \underline{87.19}          & \underline{4.13}          & \textbf{77.41} & \textbf{72.11} & \underline{87.38}          & 4.00             & 72.59          & 76.70           & 82.19          & 3.98          & 69.36          & 70.99          \\ \midrule[1pt]
\modi{DeepTrans-7B (Cold Start)} & \modi{85.06} & \modi{3.94} & \modi{66.72} & \modi{71.49} & \modi{85.90} & \modi{4.03} & \modi{69.01} & \modi{76.91} & \modi{85.22} & \modi{3.97} & \modi{73.61} & \modi{76.54} \\
\modi{DeepTrans-7B (Direct RL)} & \modi{87.13} & \modi{4.11} & \modi{71.43} & \modi{70.68} & \modi{86.37} & \modi{4.01} & \modi{72.93} & \modi{75.97} & \modi{\underline{85.92}} & \modi{4.09} & \modi{74.14} & \modi{76.00} \\
DeepTrans-7B$^{\diamondsuit}$ (Our)              & \textbf{88.84}$^{\dagger}$ & \textbf{4.21}$^{\ddagger}$ & \underline{75.38}          & 71.82          & \textbf{87.95}$^{\dagger}$ & \textbf{4.22}$^{\dagger}$ & \textbf{76.92}$^{\dagger}$ & 77.04          & \textbf{87.95}$^{\dagger}$ & \textbf{4.22}$^{\dagger}$ & \textbf{76.92}$^{\dagger}$ & \underline{76.65} \\ \bottomrule[1pt]
\end{tabular}
}
\caption{Comparison results of DeepTrans and open-source baselines (< 30B parameters). The \textbf{bold} and the \underline{underline} denote the best and second-best scores, respectively. ``$\dagger$'' and``$\ddagger$'' denote statistically significant better than the DRT-14B with t-test p < 0.01 and 0.05, respectively. ``$\diamondsuit$'' denotes models are trained on MetaphorTrans.}
\label{table:main_res_with_baselines_group1}
\end{table*}

\subsection{Implementation Details.}

\noindent \textbf{Cold Start SFT.}
We randomly select 4K English sentences from WMT24\footnote{\url{https://www2.statmt.org/wmt24/index.html}} in the general domain.
Then, DeepSeek-R1 is employed to translate these sentences from English to Chinese in a deep reasoning manner.
The synthesized 4K samples are used to SFT DeepTrans, named, cold-start SFT.
Llama-Factory framework~\cite{zheng-etal-2024-llamafactory} is used during the SFT stage.
We conduct experiments on 8$\times$NVIDIA H20 GPUs (96G) with 1e-5 learning rate and 8 (8$\times$1) batch size.
DeepSpeed ZeRO-3 optimization~\cite{rasley2020deepspeed} is also used during SFT.
We set the number of SFT epochs to 2, and it costs about 1 GPU hour.

\vspace{0.5ex}
\noindent \textbf{RL Training.}
We use GRPO RL algorithm implemented by verl\footnote{\url{https://github.com/volcengine/verl}}.
2$\times$8 H20 GPUs are used, where 8 GPUs are used to deploy DeepSeek-v3 (awq quantization) as the reward model, and another 8 GPUs are used to optimize the policy model.
We set the batch size to 64, the learning rate to 1e-6, the rollout number to 8 and the rollout temperature to 0.6, and the KL loss coefficient to 1e-3.
The number of training epochs is set to 2.
\modi{Since the scales of $r_{\text{trans}}$ and $r_{\text{thought}}$ are different, we set the trade-off hyperparameter $\alpha$ in Eq.~\ref{eq:4} to 20.}
The RL training costs 2K GPU hours.

\begin{table*}[t]
\centering
\resizebox{0.98\textwidth}{!}
{
\begin{tabular}{lcccccccccccc}
\toprule[1pt]
\multicolumn{1}{c}{\multirow{2}{*}{Model}}      & \multicolumn{4}{c}{MetaphorTrans}                                & \multicolumn{4}{c}{\emph{O. Henry}}                                     & \multicolumn{4}{c}{\emph{Orbital}}                                      \\
\cmidrule(r){2-5} \cmidrule(r){6-9} \cmidrule(r){10-13} & GRF            & GEA5          & GEA100         & CometKiwi      & GRF            & GEA5          & GEA100         & CometKiwi      & GRF            & GEA5          & GEA100         & CometKiwi      \\ \midrule[1pt]
GPT-4o                    & 85.57 & 3.86 & 71.88  & \underline{73.01}     & 88.30  & 4.00    & 71.06  & 76.74     & 85.91 & 4.17 & 73.54  & 77.67     \\
o1-preview                & 87.11 & 4.06 & \textbf{78.01}  & \textbf{73.70}      & \underline{89.73} & 4.14 & 76.17  & \textbf{78.41}     & 86.85 & \underline{4.26} & 76.80   & \textbf{78.86}     \\
QwQ-32B-preview           & 86.31 & 4.00    & \underline{75.50}   & 71.48     & 87.61 & 4.03 & 70.79  & 76.86     & 84.79 & 4.04 & 71.03  & 76.17     \\
QwQ-32B                   & \underline{88.06} & \underline{4.09} & 74.38  & 72.88     & 88.02 & \underline{4.21} & 76.36  & \underline{77.71}     & \underline{87.83} & 4.15 & 76.55  & 77.55     \\
DeepSeek-Qwen-32B         & 84.78 & 3.87 & 71.88  & 71.93     & 87.03 & 4.03 & 70.81  & 76.75     & 85.36 & 4.16 & 73.62  & \underline{77.80}      \\
DeepSeek-R1               & 84.29 & 4.02 & 73.78  & 68.33     & \textbf{89.79} & 4.17 & \textbf{77.03}  & 77.01     & 87.37 & \textbf{4.27} & \textbf{80.06}  & 76.17     \\ \midrule[1pt]
DeepTrans-7B (Our)              & \textbf{88.84} & \textbf{4.21} & 75.38  & 71.82     & 87.95 & \textbf{4.22} & \underline{76.92}  & 77.04     & \textbf{87.95} & 4.22 & \underline{76.92}  & 76.65   \\ \bottomrule[1pt]
\end{tabular}
}
\caption{Comparison results of DeepTrans and baselines (> 30B parameters or commercial LLMs).}
\label{table:main_res_with_baselines_group2}
\end{table*}

\subsection{Baselines}

\noindent \emph{(1) General Non-reasoning LLMs.}
We leverage Llama-3.1-8B-Instruct\footnote{\url{https://huggingface.co/meta-llama/Llama-3.1-8B-Instruct}}~\cite{grattafiori2024llama}, Qwen2.5-7B-Instruct\footnote{\url{https://huggingface.co/Qwen/Qwen2.5-7B-Instruct}}, Qwen2.5-14B-Instruct\footnote{\url{https://huggingface.co/Qwen/Qwen2.5-14B-Instruct}}~\cite{yang2024qwen2} and GPT-4o~\cite{hurst2024gpt} as baselines.

\noindent \emph{(2) General Reasoning LLMs.}
QwQ-32B-preview\footnote{\url{https://huggingface.co/Qwen/QwQ-32B-Preview}}, QwQ-32B\footnote{\url{https://huggingface.co/Qwen/QwQ-32B}}~\cite{team2024qwq}, Marco-o1-7B\footnote{\url{https://huggingface.co/AIDC-AI/Marco-o1}}~\cite{zhao2024marco}, DeepSeek-Qwen-7B\footnote{\url{https://huggingface.co/deepseek-ai/DeepSeek-R1-Distill-Qwen-7B}}, DeepSeek-Llama-8B\footnote{\url{https://huggingface.co/deepseek-ai/DeepSeek-R1-Distill-Llama-8B}}, DeepSeek-Qwen-14B\footnote{\url{https://huggingface.co/deepseek-ai/DeepSeek-R1-Distill-Qwen-14B}}, DeepSeek-Qwen-32B\footnote{\url{https://huggingface.co/deepseek-ai/DeepSeek-R1-Distill-Qwen-32B}}, DeepSeek-R1\footnote{\url{https://huggingface.co/deepseek-ai/DeepSeek-R1}}~\cite{guo2025deepseek} and o1-preview~\cite{openai_o1_2024} are used as baselines.

\noindent \emph{(3) MT Non-reasoning LLMs.}
\citet{wang2024drt} fine-tune three LLMs with only paired sentences of the MetaphorTrans training data (without thought).
This setting allows LLMs to learn the mapping from source literature sentences to the corresponding Chinese translations directly.
The fine-tuned LLMs are denoted as Llama-3.1-8B-SFT, Qwen2.5-7B-SFT and Qwen2.5-14B-SFT.

\noindent \emph{(4) MT Reasoning LLMs.}
\citet{wang2024drt} introduce DRT-7B\footnote{\url{https://huggingface.co/Krystalan/DRT-7B}}, DRT-8B\footnote{\url{https://huggingface.co/Krystalan/DRT-8B}} and DRT-14B\footnote{\url{https://huggingface.co/Krystalan/DRT-14B}} models, which are fine-tuned on the whole MetaphorTrans training data.
Given sentences, these LLMs could first reason and then translate.

To make a deeper understanding of these baselines, we summarize their key features in Table~\ref{table:baselines}.

\subsection{Main Results}

As shown in Table~\ref{table:main_res_with_baselines_group1}, the experimental results verify the effectiveness of DeepTrans-7B.
Specifically, compared with all baselines (< 30B parameters), DeepTrans-7B achieves state-of-the-art performance in most metrics, especially GEA5 and GRF.
DeepTrans-7B outperforms its backbone (Qwen2.5-7B-Instruct) by 16.3\%, 13.8\% and 9.0\% in terms of GEA5, GEA100 and GRF on MetaphorTrans.
As for strong baselines that are fine-tuned on MetaphorTrans (denoted with ``$\diamondsuit$''), they use the source sentences and the synthesized literature translations.
In contrast, DeepTrans-7B only uses the source sentences and achieves promising results, and thus avoids the data quality issue and potential bias in synthesized data.

\modi{To figure out the effects of the cold start SFT and the RL training, we also compare DeepTrans-7B with its two variants:
(1) \emph{DeepTrans-7B (Cold Start)} is only trained via the cold start SFT and does not incorporate RL training.
In contrast, (2) \emph{DeepTrans-7B (Direct RL)} is trained solely using RL training without the cold start SFT.
The experimental results in Table~\ref{table:main_res_with_baselines_group1} show that both variants underperform the original DeepTrans-7B across all metrics, demonstrating the rationality and the effectiveness of the two-stage training process.
The cold start SFT lets the backbone LLM (\emph{i.e.}, Qwen2.5-7B-Instruct) quickly learn the deep reasoning translation task, and brings MT performance improvements to the model. Without the cold start SFT, DeepTrans-7B (Direct RL) only achieves sub-optimal performance.
After the cold start SFT, the RL training stage further guides the model to acquire effective translation strategies.
The substantial improvements of DeepTrans-7B compared to DeepTrans-7B (Cold Start) verify the superiority of the RL training.
}

\begin{table}[t]
\centering
\resizebox{0.40\textwidth}{!}
{
\begin{tabular}{lccc}
\toprule[1pt]
\multicolumn{1}{c}{Model} & Flu.   & Sem.   & Lit.   \\ \midrule[1pt]
\multicolumn{4}{c}{\emph{MetaphorTrans}}        \\
Qwen2.5-14B-SFT             & 0.044     & -0.009    & -0.105    \\
DRT-14B                     & 0.079     & 0.079     & 0.109     \\
DeepTrans-7B (Cold Start)   & -0.258    & -0.291    & -0.269    \\
DeepTrans-7B                & \textbf{0.135}     & \textbf{0.221}     & \textbf{0.265}     \\ \midrule[1pt]
\multicolumn{4}{c}{\emph{O. Henry}}        \\
Qwen2.5-14B-SFT             & -0.020    & -0.019    & -0.170 \\
DRT-14B                     & 0.088     & 0.072     & 0.129 \\
DeepTrans-7B (Cold Start)   & -0.245    & -0.294    & -0.241 \\
DeepTrans-7B                & \textbf{0.177}     & \textbf{0.241}     & \textbf{0.282} \\ \midrule[1pt]
\multicolumn{4}{c}{\emph{Orbital}}        \\
Qwen2.5-14B-SFT             & -0.010    & 0.001     & -0.120    \\
DRT-14B                     & 0.091     & 0.082     & 0.109     \\
DeepTrans-7B (Cold Start)   & -0.238    & -0.274    & -0.241    \\
DeepTrans-7B                & \textbf{0.157}     & \textbf{0.191}     & \textbf{0.252}     \\ \bottomrule[1pt]
\end{tabular}
}
\caption{\modi{Human evaluation results in terms of fluency, semantic accuracy and literary quality.}}
\label{table:human_eval}
\end{table}

We also compare DeepTrans-7B with baselines (> 30B or commercial LLMs). As shown in Table~\ref{table:main_res_with_baselines_group2}, DeepTrans-7B generally outperforms QwQ-32B-preview and DeepSeek-Qwen-32B, and achieves competitive results with QwQ-32B, DeepSeek-R1, GPT-4o and o1-preview.
This result demonstrates the superiority of DeepTrans-7B, with only 7B parameters, which shows strong performance in deep reasoning MT.

\begin{figure*}[t]
\centering
\subfigure[Overall rewards ($r_\text{all}$)]{
  \includegraphics[width=0.31\linewidth]{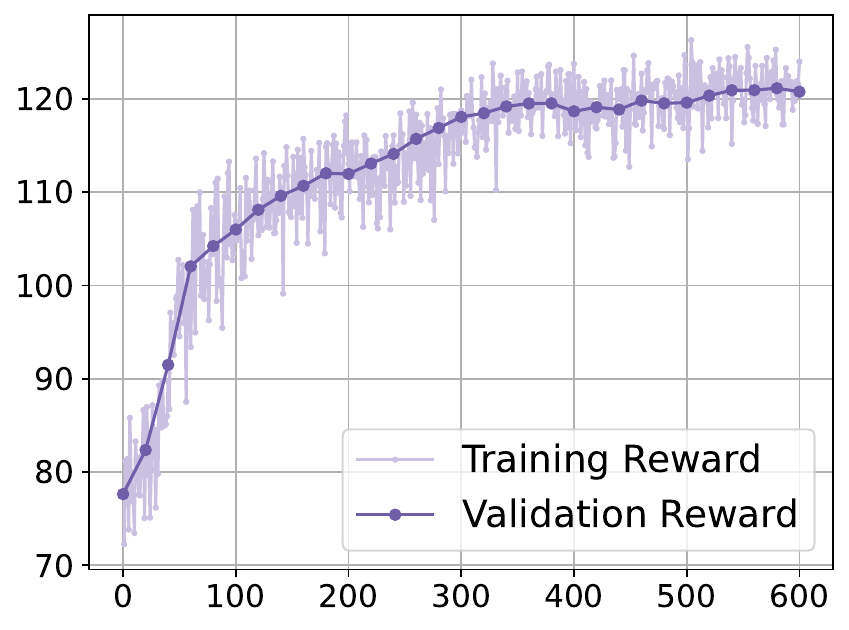}
}
\subfigure[Thought length (token-level)]{
  \includegraphics[width=0.31\linewidth]{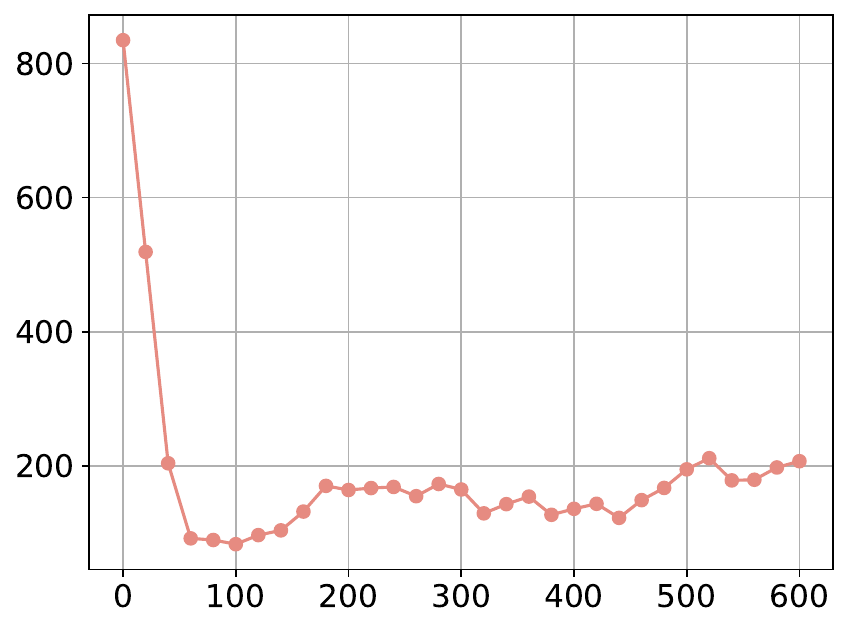}
}
\subfigure[CometKiwi]{
  \includegraphics[width=0.31\linewidth]{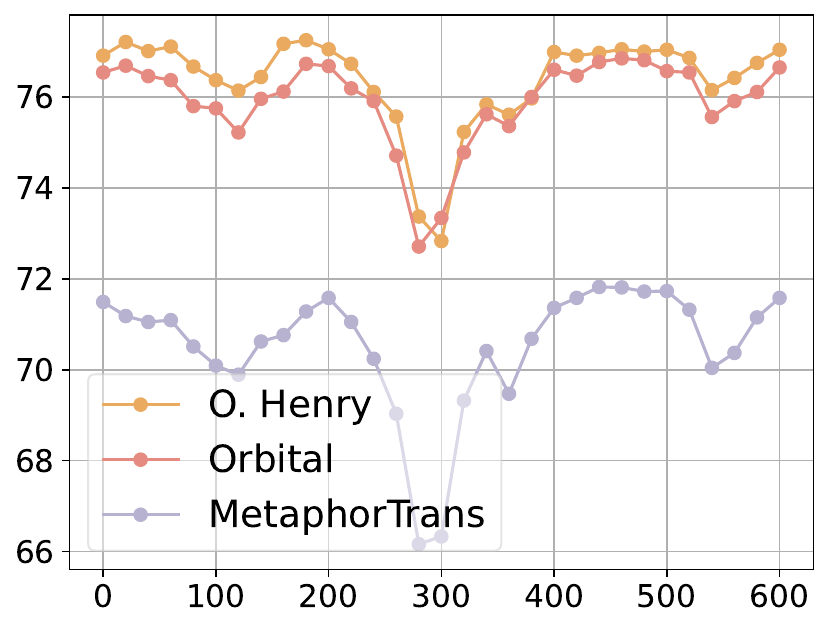}
}
\subfigure[GRF]{
  \includegraphics[width=0.31\linewidth]{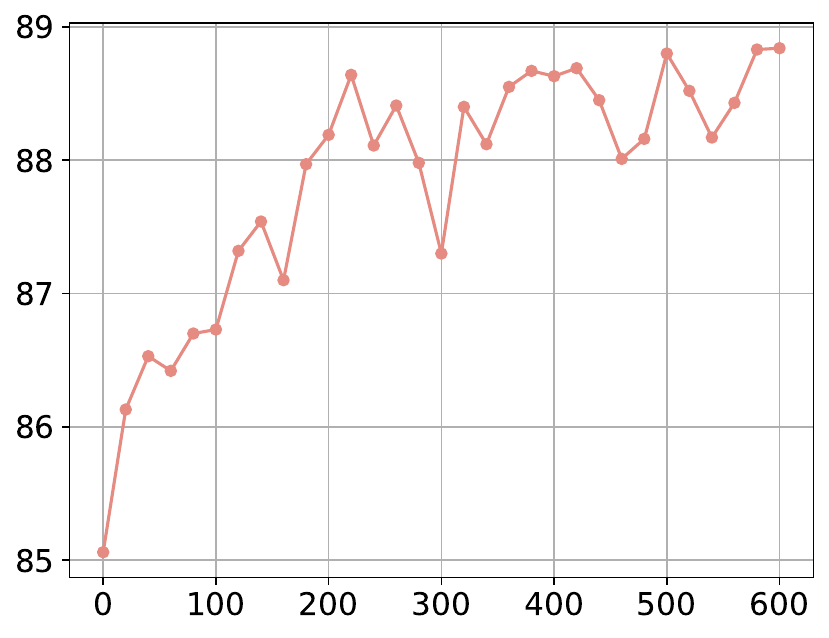}
}
\subfigure[GEA5]{
  \includegraphics[width=0.31\linewidth]{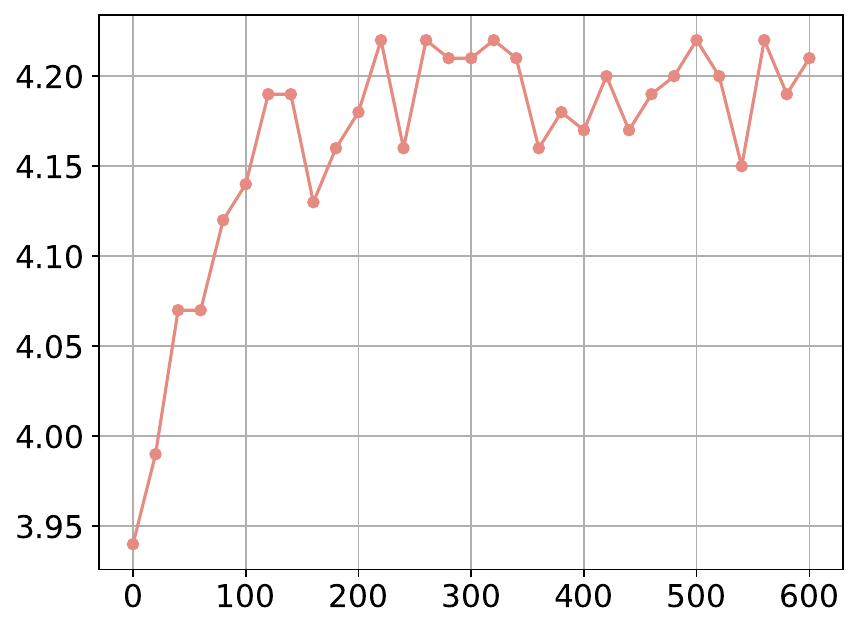}
}
\subfigure[GEA100]{
  \includegraphics[width=0.31\linewidth]{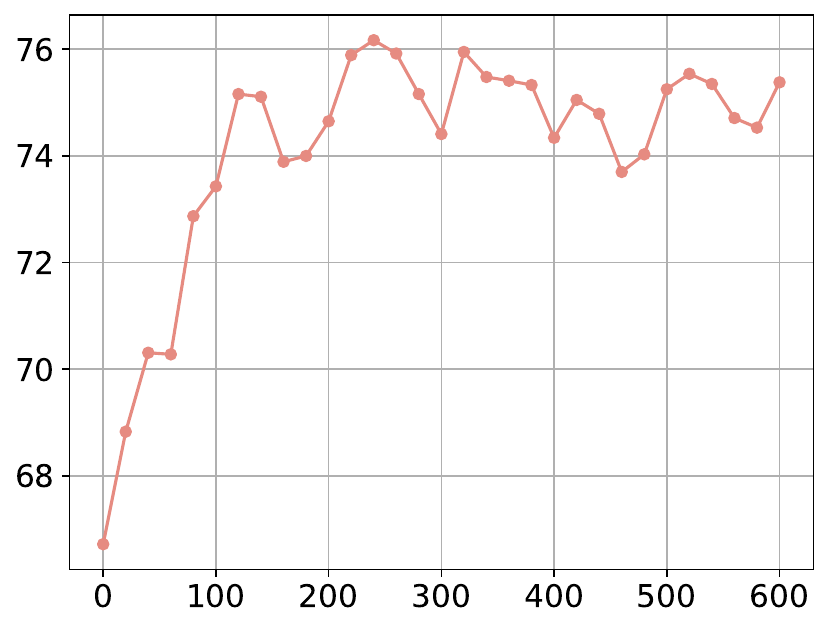}
}
\caption{Performance changes of DeepTrans-7B during RL training. Among them, (c) is conducted on all three test sets, while others are conducted on MetaphorTrans. The horizontal axis denotes the number of training steps, and there are 600 steps (2 epochs) in total. The vertical axis denotes the value of the corresponding metrics.}
\label{fig:intermediate_analyses}
\end{figure*}

\begin{table*}[t]
\centering
\resizebox{0.98\textwidth}{!}
{
\begin{tabular}{p{3.5cm}p{3.5cm}p{3.5cm}p{3.5cm}p{3.5cm}p{3.5cm}}
\toprule[1pt]
\multicolumn{1}{c}{Source Sentence}   & \multicolumn{1}{c}{DeepTrans (Cold Start)}    & \multicolumn{1}{c}{DeepTrans (150 Steps)}   & \multicolumn{1}{c}{DeepTrans (300 Steps)}     & \multicolumn{1}{c}{DeepTrans (450 Steps)}  & DeepTrans (600 Steps)   \\ \midrule[1pt]
When it was light enough Johnsy, the merciless, commanded that the shade be raised.                                                                                                   & \begin{CJK}{UTF8}{gbsn} 当光线足够明亮时，无情的约翰西斯命令把窗帘拉上。  \end{CJK}  \cometkiwi{\textbf{[CometKiwi:~81.72]}}                                & \begin{CJK}{UTF8}{gbsn} 当光线足够明亮时，那个无情的约翰西便下令拉起窗帘。  \end{CJK}  \cometkiwi{\textbf{[CometKiwi:~81.56]}}                                  & \begin{CJK}{UTF8}{gbsn} \literary{待天色稍霁}，约翰西这颗冷酷的心，便命将帘幕\literary{徐徐卷起}。   \end{CJK} \cometkiwi{\textbf{[CometKiwi:~57.61]}}       & \begin{CJK}{UTF8}{gbsn} 等到光线足够了，这个冷酷无情的约翰西，便下令将那阴影缓缓卷起。    \end{CJK}  \cometkiwi{\textbf{[CometKiwi:~72.48]}}  & \begin{CJK}{UTF8}{gbsn} 约翰西，这个冷酷无情的人，等到天色足够明亮，便命令将帘幕缓缓拉起。  \end{CJK}  \cometkiwi{\textbf{[CometKiwi:~77.28]}}  \\ \midrule
When one loves one’s Art no service seems too hard. & \begin{CJK}{UTF8}{gbsn} 当一个人热爱自己的艺术时，任何服务都显得不值一提。 \end{CJK} \cometkiwi{\textbf{[CometKiwi:~70.98]}} & \begin{CJK}{UTF8}{gbsn} 一旦钟爱艺术，便觉任何奉献皆非难事。\end{CJK} \cometkiwi{\textbf{[CometKiwi:~81.86]}} & \begin{CJK}{UTF8}{gbsn} 一旦钟爱自家的艺术，\literary{任是何般劳顿艰辛，也觉微不足道，甘之如饴。} \end{CJK} \cometkiwi{\textbf{[CometKiwi:~65.53]}} & \begin{CJK}{UTF8}{gbsn} 只要钟爱自己的艺术，再大的付出也觉甘愿。 \end{CJK} \cometkiwi{\textbf{[CometKiwi:~67.99]}} & \begin{CJK}{UTF8}{gbsn} 只要钟爱自己的艺术，再大的付出也绝不嫌苦。 \end{CJK}  \cometkiwi{\textbf{[CometKiwi:~76.16]}} \\ \midrule
For two or three days she went out canvassing for pupils. One evening she came home elated. & \begin{CJK}{UTF8}{gbsn} 连续两三天她出去招生，一个傍晚她兴高采烈地回家了。 \end{CJK} \cometkiwi{\textbf{[CometKiwi:~83.55]}} & \begin{CJK}{UTF8}{gbsn} 连续两三天，她外出物色学生。某晚归来时，她显得欣喜若狂。 \end{CJK} \cometkiwi{\textbf{[CometKiwi:~84.96]}} & \begin{CJK}{UTF8}{gbsn} 接连三四日，她外出\literary{寻觅门徒踪迹}，直至某晚归来，已得\literary{诸多佳音}满怀欣喜。 \end{CJK} \cometkiwi{\textbf{[CometKiwi:~64.59]}} & \begin{CJK}{UTF8}{gbsn} 连续一两天，她便出门去物色学生。一天傍晚，她兴冲冲地回到了家中。 \end{CJK} \cometkiwi{\textbf{[CometKiwi:~82.87]}} & \begin{CJK}{UTF8}{gbsn} 连续两三天，她外出寻觅学生上门求教。一天傍晚，她满面春风地回到了家中。 \end{CJK} \cometkiwi{\textbf{[CometKiwi:~81.21]}} \\  \bottomrule[1pt]       
\end{tabular}
}
\caption{Case Studies on \emph{O. Henry}. \literary{Purple} indicates the translations are in strong Chinese classical literary style.}
\label{table:case_study}
\end{table*}

\subsection{\modi{Human Evaluation}}
\label{subsec:he}

\modi{We conduct human evaluation to further evaluate the performance of DeepTrans-7B, DeepTrans-7B (Cold Start), DRT-14B and Qwen2.5-14B-SFT.
We randomly select 200 samples from each test set, and employ five human evaluators with high levels of fluency in English and Chinese to assess the generated translations.
The human evaluation focuses on three key aspects: fluency (Flu.), semantic accuracy (Sem.) and literary quality (Lit.).
Following~\citet{kiritchenko-mohammad-2017-best} and \citet{wang2024drt}, evaluators are tasked with identifying the best and worst translations for each aspect.
The result scores are calculated based on the percentage of times each model is selected as best minus the times it is selected as worst.
Consequently, the final scores range from -1 (indicating the worst performance) to 1 (indicating the best performance).
As shown in Table~\ref{table:human_eval}, DeepTrans-7B significantly outperforms the others, especially in literary quality.
This result demonstrates the superiority and effectiveness of DeepTrans-7B.
The Fleiss’ Kappa scores~\cite{fleiss1971measuring} of Flu., Sem. and
Lit. are 0.68, 0.70 and 0.74, respectively, indicating a good inter-agreement among evaluators.
}

\subsection{Intermediate-Stage Analyses}
\label{subsec:3.5}

To provide a deeper analysis of DeepTrans, we discuss the performance changes during RL training.
Figure~\ref{fig:intermediate_analyses} shows the corresponding details, which we analyze from the following aspects:

In terms of \emph{overall rewards} (c.f. Figure~\ref{fig:intermediate_analyses} (a)), the training rewards and validation rewards generally increase along with the training process.
The full score of the overall reward is 140 (according to Eq.~\ref{eq:4}), and DeepTrans-7B finally reaches about 120, which means $r_{\text{trans}}$ reaches at least 80. According to the definition of $r_{\text{trans}}$, 80 points indicate the translations are between ``very good translation'' and ``excellent translation'', showing the superiority of DeepTrans-7B.
Similarly, as shown in Figure~\ref{fig:intermediate_analyses} (d), (e) and (f), model performance in GRF, GEA5 and GEA100 also generally increases along with the training process.
After RL training, DeepTrans-7B demonstrates a substantial performance enhancement compared to the initial model, which is only trained via the cold start SFT. This also verifies the effectiveness of RL training.

In terms of \emph{thought length} (c.f. Figure~\ref{fig:intermediate_analyses} (b)), the average length first significantly decreases from 800+ to 100 tokens, and then slowly increases to 200 tokens.
This finding indicates that DeepTrans-7B first generates plenty of thought content owing to the cold start SFT, and thus can obtain high thought rewards.
During RL training, the model first preliminarily focuses on the translation results instead of the thought content, and it starts to simplify the thought content, since no additional thought rewards could be given.
However, when the thought length is below a threshold, the model cannot provide a meaningful thought content with short words, and it will be punished by $r_{\text{thought}}$.
Then, the model will stop simplifying the thought content, and learn to generate better translations and keep high-quality thought processes simultaneously.
Later, along with RL training, the model will find deeper thoughts that will lead to better translations, and learn to improve its thinking.

In terms of \emph{CometKiwi} (c.f. Figure~\ref{fig:intermediate_analyses} (c)), we find an interesting phenomenon: the performance drops a lot at first, and finally returns to the initial performance.
To understand the reason behind this, we provide several cases generated by the intermediate-stage DeepTrans-series models.
As shown in Table~\ref{table:case_study}, we find that the model trained with 300 steps (\emph{i.e.}, 1 epoch) has a special characteristic: \textbf{its translations exhibit a distinctly classical Chinese literary style} (\emph{e.g.}, ancient poetry).
We mark several classical-style terms in \literary{purple}, and these terms are also not commonly used in modern Chinese.
The classical style might not be well captured by the CometKiwi.
However, it still can be recognized by the GPT-4o evaluators.
As shown in Figure~\ref{fig:intermediate_analyses}, DeepTrans (300 steps) still outperforms DeepTrans (cold start) in terms of GRF, GEA5 and GEA100.
For this stylization phenomenon, we also want to discuss the following questions:
\emph{(1) Is stylization better or not?}
Though classical-style translations will provide some interesting results, we think this might lose the generalizability of the MT models.
The classical-style translations are not suitable for all genres in literature.
Our goal is to make the model provide translations that should be more easily absorbed by native people in target languages.
The classical-style translations do not follow this goal, and it should be considered a bias or preference.
Thus, we recommend that future work should also evaluate the stylization phenomenon during RL training.
\emph{(2) Why does DeepTrans first drop into the stylization phenomenon and finally get out of it?}
First, the model recognizes that the classical-style translations will receive a high translation reward, and thus, the model starts to provide classical-style translations.
This also indicates a bias/preference of the reward model.
However, such a stylization strategy will trap the model into a local optimal solution.
Fortunately, the model discovers other better solutions during exploration in RL, and finally discards the stylization strategy.
Another important question is naturally raised: \emph{which factors lead the model to get out of the local optimal solution?}
We will discuss it in \S~\ref{subsec:4.3}.

\begin{figure*}[t]
\centering
\subfigure{
  \includegraphics[width=0.35\linewidth]{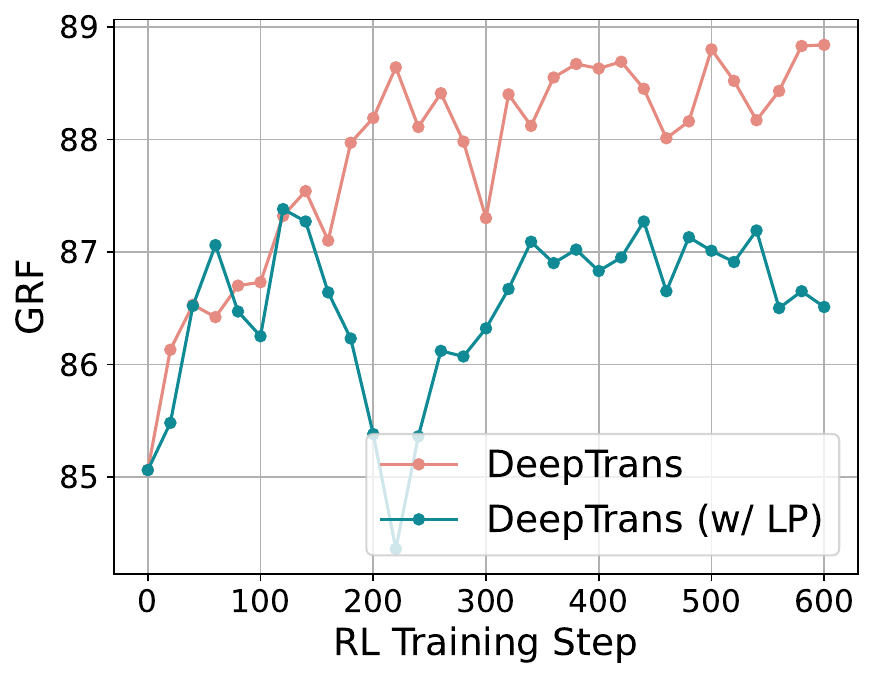}
}
\subfigure{
  \includegraphics[width=0.35\linewidth]{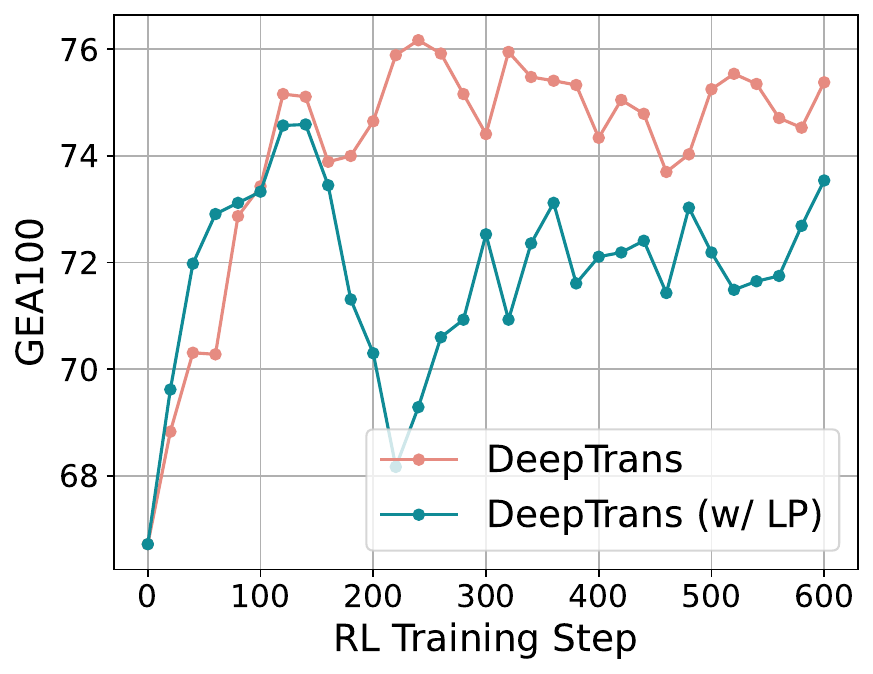}
}
\caption{The comparisons between DeepTrans and DeepTrans (w/ LP) on MetaphorTrans.}
\label{fig:failture_exp_1}
\end{figure*}

\begin{figure*}[t]
\centering
\subfigure[In terms of GRF]{
  \includegraphics[width=0.35\linewidth]{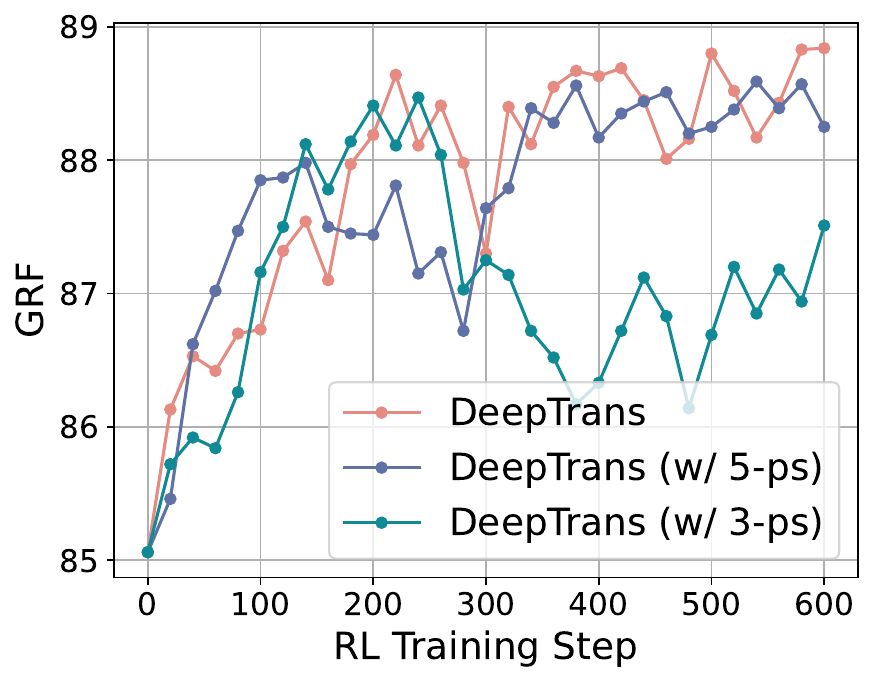}
}
\subfigure[In terms of CometKiwi]{
  \includegraphics[width=0.35\linewidth]{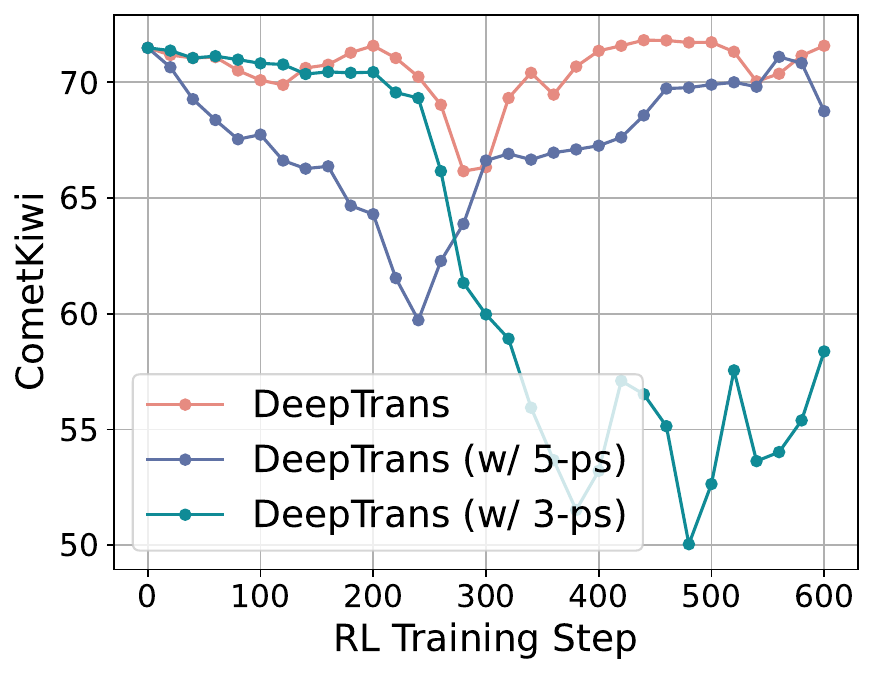}
}
\caption{Comparisons of models trained with translation rewards across different scoring scales. The results are evaluated on the MetaphorTrans test set.}
\label{fig:failture_exp_3}
\end{figure*}

\section{Failure Experience}

We discuss our failures in reward design, which could also be regarded as the ablation study of $r_{\text{all}}$.

\subsection{The Length Penalty of Thought Process.}
In the pilot RL experiments, we design a length penalty for the thought content:
\begin{equation}
\small
r_{\text{length}} = \operatorname{max} (- \frac{\operatorname{len}(\text{think})-\beta}{\eta}, 0)
\end{equation}
where $\beta$ and $\eta$ are hyperparameters. $\operatorname{len}\text{(think)}$ indicates the token-level length of the generated thought content.
In this manner, the length penalty will punish redundant thinking processes, and we add it to $r_\text{thought}$:
\begin{equation}
\small
r_\text{thought (w/ penalty)} = r_\text{thought} + r_\text{length}
\end{equation}
We set $\beta$ to 400 and $\eta$ to 400 in the pilot experiments. The DeepTrans-7B model trained with the length penalty is denoted as DeepTrans (w/ LP).
As shown in Figure~\ref{fig:failture_exp_1}, we find that the length penalty does not bring improvement during training.
In contrast, it reduces the rate of improvement in model performance.
Therefore, we do not adopt the length penalty in the final reward modeling.

\subsection{The Effect of Thought Reward}

We also attempt to remove the thought reward, and the variant overall reward is defined as:
\begin{equation}
\small
  r_{\text{all (w/o th)}} = 
\begin{cases} 
0 & \text{if  } r_{\text{format}} = 0 \\
r_{\text{trans}} & \text{if  } r_{\text{format}} \neq 0
\end{cases}
\end{equation}
However, we find that DeepTrans trained via $r_{\text{all (w/o th)}}$ quickly discards the thought process, and tends to directly output the final translation until the end of the training.
To make the model provide a valuable thought process, we further consider a short length penalty on the thought process:
If the thought length is less than $k$ tokens, we give a huge penalty to the overall reward:
\begin{equation}
\small
  r_{\text{all}\text{(w/o th \& w/ sp)}} = 
\begin{cases} 
0 & \text{if  } r_{\text{format}} = 0  \\
0 & \operatorname{len}\text{(think)} < k \\
r_{\text{trans}} & \text{else}
\end{cases}
\end{equation}
We set $k$ to 10, and find that the trained model tends to only provide several empty words in the thought content. For example, ``Well, first I have to understand the meaning of the sentence. Then, I will express it in fluent Chinese''.
This kind of thought content loses relevance and specificity towards the source sentences.

Besides, the model trained with $r_{\text{all (w/o th)}}$ or $r_{\text{all (w/o th \& w/ sp)}}$ significantly underperforms the original DeepTrans-7B in terms of all metrics.
These findings demonstrate the importance of the thought process in model application, and we should consider how to guide models to generate valuable thought during RL training.

\begin{table*}[t]
\centering
\resizebox{0.98\textwidth}{!}
{
\begin{tabular}{p{3.5cm}p{3.5cm}p{3.5cm}p{3.5cm}p{3.5cm}p{3.5cm}}
\toprule[1pt]
\multicolumn{1}{c}{\multirow{2}{*}{Source Sentence}} & \multirow{2}{*}{DeepTrans (Cold Start)} & \multicolumn{2}{c}{DeepTrans Variants (300 Steps)}                    & \multicolumn{2}{c}{DeepTrans Variants (600 Steps)}                    \\
\cmidrule(r){3-4} \cmidrule(r){5-6} &     & \multicolumn{1}{c}{w/ 3-point scale} & \multicolumn{1}{c}{w/ 5-point scale} & \multicolumn{1}{c}{w/ 3-point scale} & \multicolumn{1}{c}{w/ 5-point scale}  \\ \midrule[1pt]
When it was light enough Johnsy, the merciless, commanded that the shade be raised.                                                                                                   & \begin{CJK}{UTF8}{gbsn} 当光线足够明亮时，无情的约翰西斯命令把窗帘拉上。  \end{CJK}  \cometkiwi{\textbf{[CometKiwi:~81.72]}}                                & \begin{CJK}{UTF8}{gbsn} 当光线已然充足，约翰西，那无情之人，竟\literary{令帷幕徐徐卷起}。  \end{CJK}  \cometkiwi{\textbf{[CometKiwi:~74.46]}}                                  & \begin{CJK}{UTF8}{gbsn} 待天色渐明，那冷漠无情的约翰西，便命人将帘幕\literary{徐徐卷起}。   \end{CJK} \cometkiwi{\textbf{[CometKiwi:~75.08]}}       & \begin{CJK}{UTF8}{gbsn} 天色稍显明亮，约翰西，那颗铁石心肠，便下令将阴影卷起。   \end{CJK}  \cometkiwi{\textbf{[CometKiwi:~66.96]}}  & \begin{CJK}{UTF8}{gbsn} 待天色足够明亮，约翰西这颗铁石心肠，已然下令将帘幕卷起。  \end{CJK}  \cometkiwi{\textbf{[CometKiwi:~71.40]}}  \\ \midrule
When one loves one’s Art no service seems too hard. & \begin{CJK}{UTF8}{gbsn} 当一个人热爱自己的艺术时，任何服务都显得不值一提。 \end{CJK} \cometkiwi{\textbf{[CometKiwi:~70.98]}} & \begin{CJK}{UTF8}{gbsn} \literary{若钟情于艺道，任劳任怨亦无妨}。\end{CJK} \cometkiwi{\textbf{[CometKiwi:~69.38]}} & \begin{CJK}{UTF8}{gbsn} 一旦钟爱\literary{所钟之艺}，\literary{便觉任劳任怨非艰}。 \end{CJK} \cometkiwi{\textbf{[CometKiwi:~69.38]}} & \begin{CJK}{UTF8}{gbsn} \literary{艺海倾心无难事，任凭千般苦亦甘}。 \end{CJK} \cometkiwi{\textbf{[CometKiwi:~54.76]}} & \begin{CJK}{UTF8}{gbsn} 只要钟爱所钟情的艺术，任何付出都觉轻而易举。 \end{CJK}  \cometkiwi{\textbf{[CometKiwi:~76.59]}} \\ \midrule
For two or three days she went out canvassing for pupils. One evening she came home elated. & \begin{CJK}{UTF8}{gbsn} 连续两三天她出去招生，一个傍晚她兴高采烈地回家了。 \end{CJK} \cometkiwi{\textbf{[CometKiwi:~83.55]}} & \begin{CJK}{UTF8}{gbsn} 连续几日外出\literary{寻觅门徒}，\literary{一夕归家心欢悦}。 \end{CJK} \cometkiwi{\textbf{[CometKiwi:~80.25]}} & \begin{CJK}{UTF8}{gbsn} 她连着几日四处奔走\literary{寻觅门徒}，直至某晚，她满心欢喜地踏入了家园。 \end{CJK} \cometkiwi{\textbf{[CometKiwi:~74.54]}} & \begin{CJK}{UTF8}{gbsn} 连续几日奔走寻门徒，直至傍晚\literary{方归家}。\literary{满心欢喜步履轻，笑颜如花映晚霞}。 \end{CJK} \cometkiwi{\textbf{[CometKiwi:~55.10]}} & \begin{CJK}{UTF8}{gbsn} 她接连走了两三天，到处奔走寻觅学生。一天傍晚，她满面春风地回到了家中。 \end{CJK} \cometkiwi{\textbf{[CometKiwi:~83.87]}} \\  \bottomrule[1pt]       
\end{tabular}
}
\caption{The results of Case Studies on \emph{O. Henry}.}
\label{table:case_study2}
\end{table*}

\subsection{The Effect of Translation Reward}
\label{subsec:4.3}

We define the translation reward using a 100-point scale in \S~\ref{subsec:2.1}, and we further study the effect of the scoring criteria on translation rewards.
Specifically, we use two variants:
$r_\text{trans (3 point)}$ and $r_\text{trans (5 point)}$ simply narrow the scoring scope of $r_\text{trans}$ from a 100-point scale to a 3-point scale and a 5-point scale, respectively.
During training with these two variants, the $\alpha$ in Eq.~\ref{eq:4} is also changed to ensure the same ratio between thought rewards and translation rewards, \emph{i.e.}, we set $\alpha$ to 0.6 and 1.0 when \modi{adopting} $r_\text{trans (3 point)}$ and $r_\text{trans (5 point)}$, respectively.
We denote the models trained with the variants as ``DeepTrans (w/ 3-ps)'' and ``DeepTrans (w/ 5-ps)''\modi{, where ``ps'' is the abbreviation of ``point scale''}.
As shown in Figure~\ref{fig:failture_exp_3} (a), both variants show the sub-optimal results in terms of GRF, verifying the rationality of our original design.

As for CometKiwi in Figure~\ref{fig:failture_exp_3} (b), we find that DeepTrans (w/ 5-ps) shows similar trends with original DeepTrans, \emph{i.e.}, first drops and finally recovers.
However, DeepTrans (w/ 3-ps) continues to decline and fluctuate during training. In the end, it achieves a low CometKiwi score, and perhaps it needs more training steps to recover the performance.
In view of the stylization phenomenon we discussed in \S~\ref{subsec:3.5}, we conjecture the trends shown in Figure~\ref{fig:failture_exp_3} (b) are attributed to \emph{stylization}.

To figure it out, we provide case studies on the variant models in Table~\ref{table:case_study2}.
When the models are trained with 300 steps (1 epoch), both variants show the stylization phenomenon.
Further, when trained with 600 steps (2 epochs), we find that DeepTrans (w/ 5-ps) discards the classical-style translations. In contrast, DeepTrans (w/ 3-ps) enhances the classical-style translations.
As we also described in \S~\ref{subsec:3.5}, the basic reason of the stylization phenomenon is a preference of the reward model, which encourages the MT model to provide stylized translations as a local optimal solution.
When the reward model provides a wider scoring scope, it gives a larger latent space for the policy model to explore.
Thus, the policy model is more likely to get out of the local optimal solution.
If using a narrow reward scope, some minor changes in the policy model will not be reflected by the reward signal immediately, making it hard to transfer the translation strategies.
In conclusion, \emph{the scoring scope of the reward models has a great effect on the policy model, and it is necessary to employ a wider reward scope for the model to explore a good translation strategy.}

\section{Conclusion \modi{and Future Work}}

In this work, we propose DeepTrans, which aims to enhance LLMs' deep reasoning MT ability via RL.
We use DeepSeek-v3 as the reward model to provide discrete reward values.
The experimental results in the literature MT show the effectiveness of DeepTrans.
It outperforms QwQ-32B-preview, and achieves competitive results with state-of-the-art deep reasoning LLMs.
In addition, we give deeper analyses on model results during RL training, and find a stylization phenomenon that needs to be carefully considered.
Moreover, we summarize the failure experiences and critical components in the RL framework to provide a deeper understanding of deep reasoning MT LLMs.

\modi{In the future, the following directions may be worth exploring to promote the deep reasoning translation:
(1) Extending deep reasoning translation models into multi-lingual MT. Consequently, the model could think and translate among different languages;
(2) Qualifying and controlling the translation style during RL training to alleviate the stylization phenomenon (Section~\ref{subsec:3.5});
(3) Exploring translation rewards with enhanced accuracy, robustness, or efficiency. In this work, we prompt vanilla DeepSeek-v3 to provide the translation reward, which is computationally intensive;
(4) Balancing long- and short-thought to improve models' inference efficiency, and avoid the overthinking issue~\cite{sui2025stop}. For simple translation queries, a long thought process may not be necessary.
}

\section{\modi{Limitation}}

\modi{While we show the effectiveness of reinforcement learning in deep reasoning translation, there are some limitations worth noting:
(1) We evaluate the deep reasoning translation models in a single translation direction, \emph{i.e.}, English-to-Chinese, and future work could extend our method to other directions;
(2) The reward model employed in our experiments is DeepSeek-v3, which requires significant computational resources to infer;
(3) Due to resource limitations, we conduct experiments on LLMs with 7B parameters, and future work could extend our method to LLMs with more parameters.}

\bibliography{tacl2021}
\bibliographystyle{acl_natbib}

\iftaclpubformat

\onecolumn

\appendix

\fi

\end{document}